\documentclass[final,5p,times]{elsarticle}

\usepackage{amssymb}
\usepackage[nodots]{numcompress}

\usepackage{url}
\usepackage{amssymb}
\usepackage[cmex10]{amsmath}
\usepackage{setspace}  
\usepackage{multirow}
\usepackage{rotating}
\usepackage[table]{xcolor}
\usepackage{hhline}
\usepackage{threeparttable}
\usepackage{amsmath}
\usepackage{algorithm}
\usepackage{algpseudocode}
\usepackage{graphicx}
\usepackage{caption}
\usepackage{subcaption}
\usepackage{epstopdf}
\journal{Pattern Recognition}

\begin{document}

\begin{frontmatter}

\title{Active Scene Learning}

\author{Erelcan Yan\i k}
\ead{yanikerelcan@gmail.com}

\author{Tevfik Metin Sezgin}
\ead{mtsezgin@ku.edu.tr}
\ead[url]{http://iui.ku.edu.tr}

\address{Ko\c c University, College of Engineering, Istanbul, Turkey}

\begin{abstract}

Sketch recognition allows natural and efficient interaction in pen-based interfaces. 
A key obstacle to building accurate sketch recognizers has been the difficulty 
of creating large amounts of annotated training data. Several authors have attempted to 
address this issue by creating synthetic data, and by building tools that support efficient 
annotation. Two prominent sets of approaches stand out from the rest of the crowd. They use 
interim classifiers trained with a small set of labeled data to aid the labeling of the remainder of
the data. The first set of approaches uses a classifier trained with a partially labeled dataset
to automatically label unlabeled instances. The others, based on active learning, save 
annotation effort by giving priority to labeling informative data instances. The former
is sub-optimal since it doesn't prioritize the order of labeling to favor informative instances, while 
the latter makes the strong assumption that unlabeled data comes in an already segmented 
form (i.e. the ink in the training data is already assembled into groups forming isolated object 
instances). In this paper, we propose an active learning framework that combines the strengths
of these methods, while addressing their weaknesses. In particular, we propose two methods
for deciding how batches of unsegmented sketch scenes should be labeled. The first method, 
scene-wise selection, assesses the informativeness of each drawing (sketch scene) as a whole, and
asks the user to annotate all objects in the drawing. The latter, segment-wise selection, attempts
more precise targeting to locate informative fragments of drawings for user labeling. 
We show that both selection schemes outperform random selection. Furthermore, 
we demonstrate that precise targeting yields superior performance. Overall, 
our approach allows reaching top accuracy figures with up to 30\% savings in annotation cost.
\end{abstract}


\end{frontmatter}

\section{Introduction}

Sketch recognition has enabled computer graphics applications in a wide range of domains from education \cite{ref1,ref2} to design \cite{ref3,ref4,ref5}. A widely acknowledged problem standing in the way of building accurate sketch recognizers has been the necessity of obtaining large amounts of annotated data.

Researchers have tried to alleviate the cost of annotation through a variety of strategies. For example, there has been considerable work to build methods to learn from few examples. These include methods based on synthetic data generation \cite{ref6}, and self-learning \cite{ref7}. Unfortunately these methods fall short of delivering accuracy values that are attainable by labeling large sets of examples, hence they are sub-optimal. Others have suggested improving the user experience of data annotation through custom annotation interfaces \cite{ref8,ref9,ref10,ref11}. Yet improving the user experience does not reduce the number of items that have to be annotated. 

Two methods in the literature aim to reduce the cost of annotation by actually reducing the number of instances that the user is asked to annotate without sacrificing accuracy. The first \cite{ref12} utilizes auto-annotation while the other \cite{ref13} is based on active learning. Both methods attempt to reduce the number of required annotations with the help of classifiers trained with a partially labeled dataset. 

Auto-annotation automatically labels unlabeled data, and asks the user to correct mislabeled instances. This method requires the annotator to verify the labels for all instances. Furthermore, unlike active learning, it does not harvest the benefits of  focusing the labeling effort on the more informative unlabeled examples.

Active learning is a machine learning strategy that aims to reduce the labeling effort by selecting the most informative samples from a pool of unlabeled data. The AL process is initialized by training a classifier with a few labeled samples, the so-called ``seed set.'' This classifier is used to assign class probabilities to the unlabeled samples, which in turn is used to compute the informativeness of the unlabeled instances. The user labels the informative instances, and the classifier is gradually improved by re-training with the extended set of labeled data. The learning process continues in rounds until a target validation accuracy is achieved or until we run out of resources (e.g. labeling time or computational resources). Active learning has been shown to reduce annotation effort for segmented data \cite{ref13}. 

Unfortunately, active learning assumes that the unlabeled data has already been segmented. Extending active learning to work with unsegmented data is nontrivial. There is limited amount of work using active learning for sequence labeling and semantic segmentation. However, existing approaches assume that their input can be decomposed into primitive units (e.g. super-pixels in computer vision, tokens in text), and carry out labeling and active learning at the granularity of these units \cite{ref14,ref15,ref16,ref17,ref18,ref19,ref20,ref21,ref22,ref23,ref24}. Here, active learning proceeds at the level of primitives, essentially similar to how it would in the isolated active learning case, usually with some additional constraints on pairwise primitives (e.g. through Markov Random Fields or n-gram representations). 

In this paper, we present a new active learning framework, Active Scene Learning (ASL), which extends the state of the art to handle unsegmented sketch scenes at the granularity of entire objects. Our approach is designed as a continuous cycle consisting of segmentation and annotation steps. 

In the segmentation step, the unlabeled scenes in the dataset are segmented using classifiers trained on partially labeled data. This yields a set of isolated objects. We call these objects {\em candidate objects}, because the user is eventually asked to label instances chosen among these candidates based on their informativeness. Note that the candidate objects are extracted using automated segmentation, hence the predicted boundaries, as well as the class labels, may be erroneous. 

The annotation step takes the output of the segmentation process, and identifies entire scenes or parts of scenes to be labeled by the user. We refer to these approaches as {\em scene-wise selection} strategy, which prioritizes labeling informative scenes; and {\em segment-wise selection} strategy, which prioritizes labeling informative scene segments.

In this paper, we demonstrate that Active Scene Learning helps annotation using sketch recognition as a case study. Our results show that Active Scene Learning is effective in reducing the cost of annotation. The results and the approach presented in this paper open the avenue for applying active learning to unsegmented data, and is of interest to practitioners of sketch recognition as well as a broader community of practitioners who rely on machine learning in their applications. Specifically, our main contributions can be listed as:

\begin{itemize}
	\item We propose the Active Scene Learning (ASL) framework to enable AL on scene data.
    \item We show that candidate objects carry valuable information which can be utilized to guide AL even if they do not have accurate segmentation boundaries.
    \item We propose two selection schemes under the ASL framework, scene-wise selection and segment-wise selection, and demonstrate that both schemes outperform random selection.
    \item We demonstrate that precise-targeting with segment-wise selection yields superior performance in comparison to scene-wise selection.
    \item We show that ASL can reach top-accuracy figures with up to 30\% savings in annotation cost.
\end{itemize}

This paper is organized as follows: First, we introduce the ASL framework and describe the scene-wise and segment-wise selection schemes. In Section 3, we first describe the datasets used in our experiments, then the details of our experimental design. In Section 4, we describe the evaluation metric employed in our analysis and then present the analysis methodology. We present the analysis results with a discussion in Section 5. Finally, we conclude with related work and a summary of future research directions.

\section{Active Scene Learning}

Active Scene Learning (ASL) is a framework aiming to reduce annotation cost for a pool of unlabeled scenes by exploiting active learning. A scene is composed of an arbitrary number of domain objects in arbitrary configurations. The key idea behind ASL is to obtain annotation gains, when building a segmentation solution, by utilizing information on candidate objects extracted via a segmentation algorithm.

\subsection{Preliminaries and Definitions}

In the sketching domain, a scene consists of a collection of strokes. To ease the segmentation process, we break the strokes into basic geometric primitives (e.g. arcs, lines etc.). The primitives serve as the smallest unit of segmentation. At this point, a sketch scene becomes a collection of primitives. The goal of active scene learning is defined as grouping these primitives (to form objects) accurately and assigning labels to the groups (of primitives), with minimal user (annotator) effort. Consider the example sketch scene in Figure 1a. It consists of 3 objects, 12 strokes and 17 primitives. Figure 1b shows four of the several possible segmentations along with their candidate objects, which may or may not match the ground truth. The mismatch may be a result of incorrect segmentation or misclassification of a correctly segmented group of primitives.

\begin{figure*}[!htbp]
        \centering
        \begin{subfigure}[b]{0.45\textwidth}
                \includegraphics[height=0.65\textwidth, width=\textwidth]{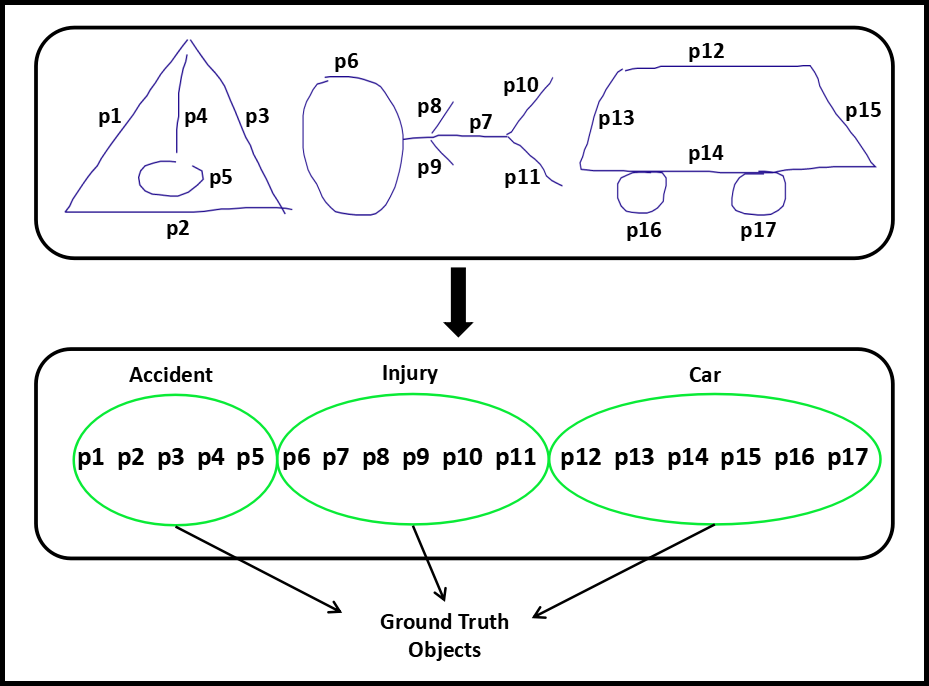}
                \caption{An example scene and its ground truth objects.\newline}
                \label{fig:Figure1a}
        \end{subfigure}
        \hspace{0.08\textwidth}
        \begin{subfigure}[b]{0.45\textwidth}
                \includegraphics[height=0.65\textwidth, width=\textwidth]{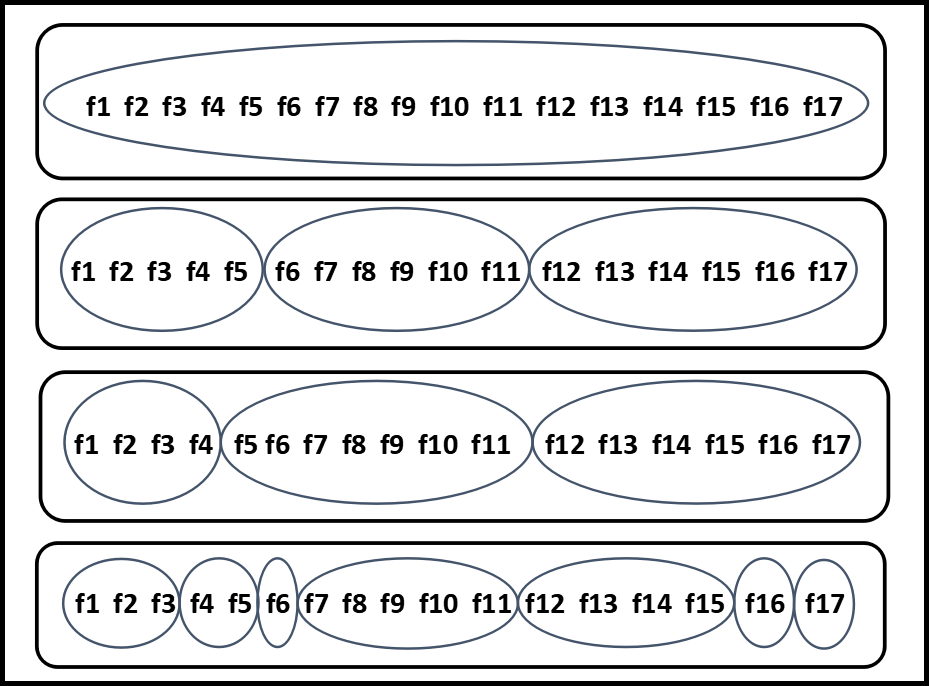}
                \caption{Four of several possible segmentations of the scene, described by grouping primitives.}
                \label{fig:Figure1b}
        \end{subfigure}
        \caption{An illustration of a scene composed of its primitives (in this case, lines and arcs). An object constitutes to a group of primitives. A candidate object may or may not match the ground truth object depending on the accuracy of the segmentation model.}\label{fig:scene}
\end{figure*}

A scene $S$ is composed of primitives $p$, and groups of primitives ($p$) constitute objects $O_i$. $O_i$ are ground truth objects while $C_i$ are candidate objects. Let $S_i=(O_1, O_2, ..., O_n)$ where $n$ is the number of ground truth objects in the scene. Also, let $S_i=(p_1, p_2, ..., p_k)$ where k is the number of primitives in the scene. Assuming $O_i=\{p_j, p_{j+1}, ..., p_{j+r}\}$ and $C_i=\{p_h, p_{h+1}, ..., p_{h+q}\}$, then $C_i$ constitutes to a correctly segmented candidate iff $j=h$ and $r=q$. In other words, if a candidate object (a group of primitives) extracted by the segmentation algorithm matches each primitive of any ground truth object in the scene, we say that it is a correctly segmented candidate object; otherwise call it a mis-segmented candidate object.

In the following subsections, we present two ASL approaches: scene-wise selection and segment-wise selection.

\begin{algorithm*}[!htbp]
  \caption{Scene-wise selection}
  \begin{algorithmic}[1]    
    \State Initialize the model with the seed set (a set of labeled isolated objects).
    \Repeat
    	\State Segment and recognize the scenes in the pool with the current model.
    	\State \parbox[t]{\dimexpr\linewidth-\algorithmicindent}{Compute scene informativeness scores with a desired scene-wise selection method.\strut}
    	\Repeat
    	\State Select the most informative unlabeled scene from the pool.
    		\State \parbox[t]{\dimexpr\linewidth-\algorithmicindent}{Request boundaries and labels for the ground truth objects in the scene.\strut}
    		\State Add all the objects in the scene to the training set.
    		\State Remove the scene from the pool.
    	\Until{the batch is full}
    	\State Re-train the model with the extended training set.
    \Until{the halting point is reached} \\
    \Return the model
  \end{algorithmic}
\end{algorithm*}

\begin{algorithm*}[!htbp]
  \caption{Segment-wise selection}
  \begin{algorithmic}[1]
    \State Initialize the model with the seed set.
    \Repeat
    	\State Segment and recognize the scenes in the pool with the current model.
    	\State Compute informativeness score for each candidate object.
    	\State Create an empty set (``processed-set'') to keep track of processed candidate objects.
    	\Repeat
    	    \State \parbox[t]{\dimexpr\linewidth-\algorithmicindent}{Select the highest scoring candidate object which is not in the ``processed-set''}
    		\State \parbox[t]{\dimexpr\linewidth-\algorithmicindent}{Request boundaries and labels for the ground truth objects intersecting with the current candidate object.\strut}
    		\State Add the ground truth objects to the training set if they are not already added (check though less likely).
    		\State \parbox[t]{\dimexpr\linewidth-\algorithmicindent}{Add the current candidate object to the ``processed-set''.\strut}
    	\Until{the batch is full}
    	\State Re-train the model with the extended training set.
    \Until{the halting point is reached} \\
    \Return the model
  \end{algorithmic}
\end{algorithm*}

\subsection{Scene-wise Selection}

Scene-wise selection aims to assign an informativeness score to each unlabeled scene in the pool after segmenting and recognizing the contents. Then, it extends the training data with all the objects in the selected scenes while using the boundaries and labels for the ground truth objects in these scenes provided by the annotator. Algorithm 1 depicts the generic methodology for scene-wise selection. We have devised four methods for computing a scene-wise informativeness score, described below.

\subsubsection{Arithmetic Mean based Aggregation (ArM)}

ArM is a simple method that combines informativeness scores of candidate objects. Let ${C_k}$ to be the set of candidate objects extracted from a scene $S_i$ with length $K$. Let $I_k$ be the informativeness score computed for the candidate object $C_k$. Then, the scene score $SS_i$ is the arithmetic mean of the scores of the candidate objects in the scene: \[SS_i = \frac{1}{K} \sum_{k=1}^{K} I_k\]

\subsubsection{Geometric Mean based Aggregation (GM)}

GM penalizes scenes with uninformative candidate objects more than ArM. Let ${C_k}$ be the set of candidate objects extracted from a scene $S_i$ with length $K$. Let $I_k$ be the informativeness score assigned to a candidate object $C_k$. Then, the scene score $SS_i$ is the geometric mean of the scores of the candidate objects in the scene: \[SS_i = (\prod _{i=1}^{K} I_k)^{\frac {1}{K}}\]

\subsubsection{Maximum of the Scene Aggregation (MoS)}

MoS sets the scene score to the informativeness score of the candidate object with the maximum score. We expect MoS to perform worse in the earlier rounds of AL, but allow a sharper focus for final rounds when many candidate objects are uninformative. Let ${C_k}$ be the set of candidate objects extracted from a scene $S_i$ with length $K$. Let $I_k$ be the informativeness score computed for a candidate object $C_k$. Then, the scene score $SS_i$ is defined as: \[SS_i = \max_{k=1,\dots,K} I_k\]

\subsubsection{Aggregation based Scene Probability (SP)}

SP focuses on scenes that are hardest to interpret. SP assigns scene score to one minus the probability of the most probable interpretation of the scene. Let $P(S_i)$ be the probability of the most probable interpretation of the scene $S_i$. Then, scene score $SS_i$ is: \[SS_i = 1 - P(S_i)\]

\subsection{Segment-wise Selection (SwS)}

Scene-wise selection methods have several drawbacks. A selected informative scene may contain uninformative candidate objects along with the informative ones. It is important to note that the informativeness of a particular scene is conditional on the classifier at hand. In the earlier rounds of active learning where we have a relatively weak classifier, most scenes are likely to achieve high informativeness scores, and, if labeled, would improve the recognizer at hand. In the later rounds, however, scenes become less informative. Furthermore, informative scenes tend to contain a large fraction of uninformative objects. Hence, it becomes imperative to identify the relatively more informative scenes among many uninformative ones, and furthermore identify specific regions within the scenes that are likely to contain the objects/regions that make the scenes informative. Therefore, we propose segment-wise selection strategy, which is depicted in Algorithm 2, to avoid such uninformative candidate objects.

Segment-wise selection (SwS) attempts precise targeting to locate informative objects within the scenes. In particular, it considers the candidate objects that the segmentation model finds most difficult to segment and recognize. Then, it requests boundaries and labels for the ground truth objects intersecting the boundaries of the candidate object. Hence, SwS focuses on scenes locally rather than asking annotation for all the objects in the scene. Note that the candidate objects are still extracted through global segmentation. Consequently, the informativeness scores are based on the results of the global segmentation.

\section{Experimental Design}

We first describe the datasets and the segmentation algorithm employed in our experiments, and then explain the structure of our experimental design in this section.

\subsection{Datasets}

We conduct experiments on sketch scenes constructed from the COAD and the NicIcon datasets. The COAD database contains a total of 620 samples from 20 different symbol classes, whereas the NicIcon database contains a total of 22958 samples from 14 different classes.

We have a multi-phase scene generation approach, presented in Algorithm 3, to obtain data matching our experimental design, which we elaborate on in Section 4. First, we prepare folds, the so-called ``combined-folds'', containing shuffled isolated objects from each class. Each class contributes samples to combined-folds in a balanced manner. Then, we obtain source training and source test sets for each \textit{Repeat} by shifting and grouping these combined-folds. Finally, we create scenes (for training and test) by sampling isolated objects from source training and source test sets.

\begin{algorithm*}[!htbp]
  \caption{Synthetic Scene Generation}
  \begin{algorithmic}[1]
    \State Create N equally sized folds for each class by random sampling (see Figure 2a).
    \State Merge each respective fold from each class to obtain combined-folds (see Figure 2b).
    \For {N \textit{Repeats} (See Figure 3)}
    	\State Group isolated objects of first N-1 combined-folds to obtain source training set for current \textit{Repeat}.
    	\State Assign N\textsuperscript{th} combined-fold as source test set for current \textit{Repeat}.
    	\State Circular-shift combined-folds.
    \EndFor
    \For {N \textit{Repeats}}
    	\State Sample and remove S seeds from source training set.
    	\State Add seeds to seed set of current \textit{Repeat}.
    \EndFor
    \For {Each source training/test set}
    	\For {Each scene-size option k}
    		\Repeat
    			\State Randomly select k classes
    			\State Sample and remove an isolated object belonging to each class from corresponding source training/test set.
    			\State Compose a scene by utilizing these samples and add it to corresponding training/test scene set.
    		\Until desired number of scenes created for size k
    	\EndFor
    \EndFor \\
    \Return seed set, training scenes and test scenes for each \textit{Repeat}
  \end{algorithmic}
\end{algorithm*}

\begin{figure*}[!htbp]
        \centering
        \begin{subfigure}[b]{0.35\textwidth}
                \includegraphics[height=0.65\textwidth, width=\textwidth]{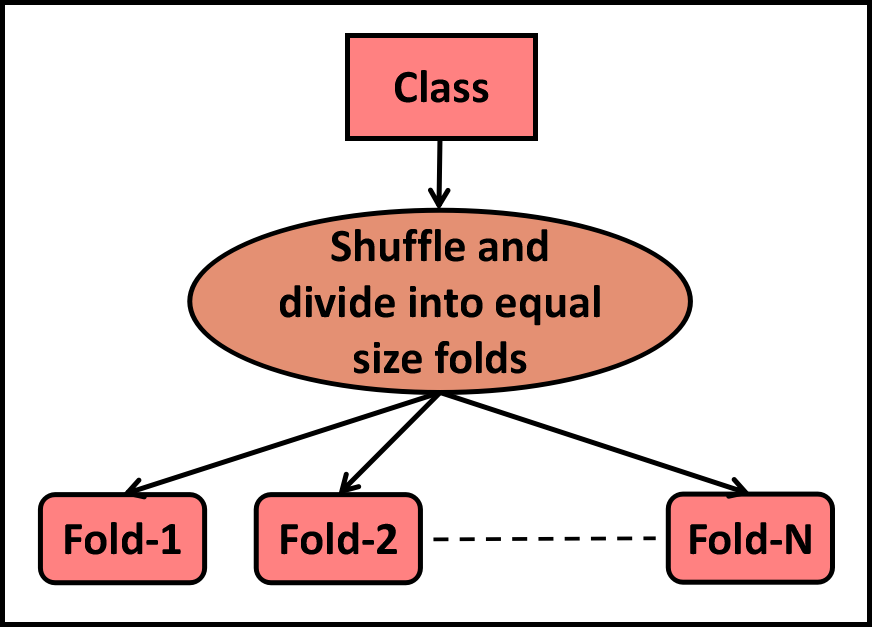}
                \caption{Create folds from samples of a class.}
                \label{fig:Figure2a}
        \end{subfigure}
        \hspace{0.1\textwidth}
        \begin{subfigure}[b]{0.35\textwidth}
                \includegraphics[height=0.65\textwidth, width=\textwidth]{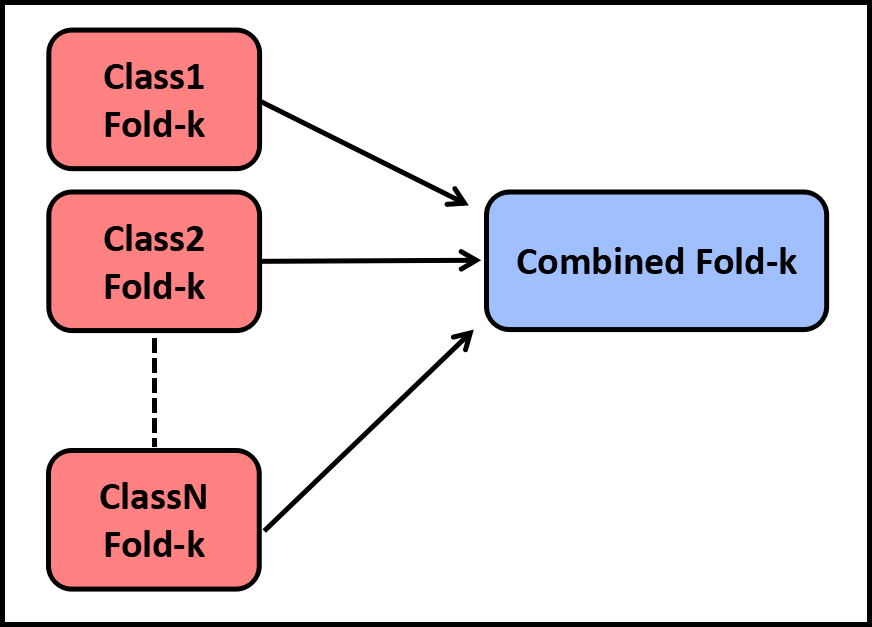}
                \caption{Combine corresponding folds of each class.}
                \label{fig:Figure2b}
        \end{subfigure}
        \caption{For each class, divide the data into N folds by random sampling. Then, merge corresponding folds of each class into the so-called ``Combined-folds''. Note that combined folds are mutally exclusive.}\label{fig:sceneCreation1}
\end{figure*}

\begin{figure*}[!htbp]
	\centering
	\includegraphics[height=0.35\textwidth, width=0.60\textwidth]{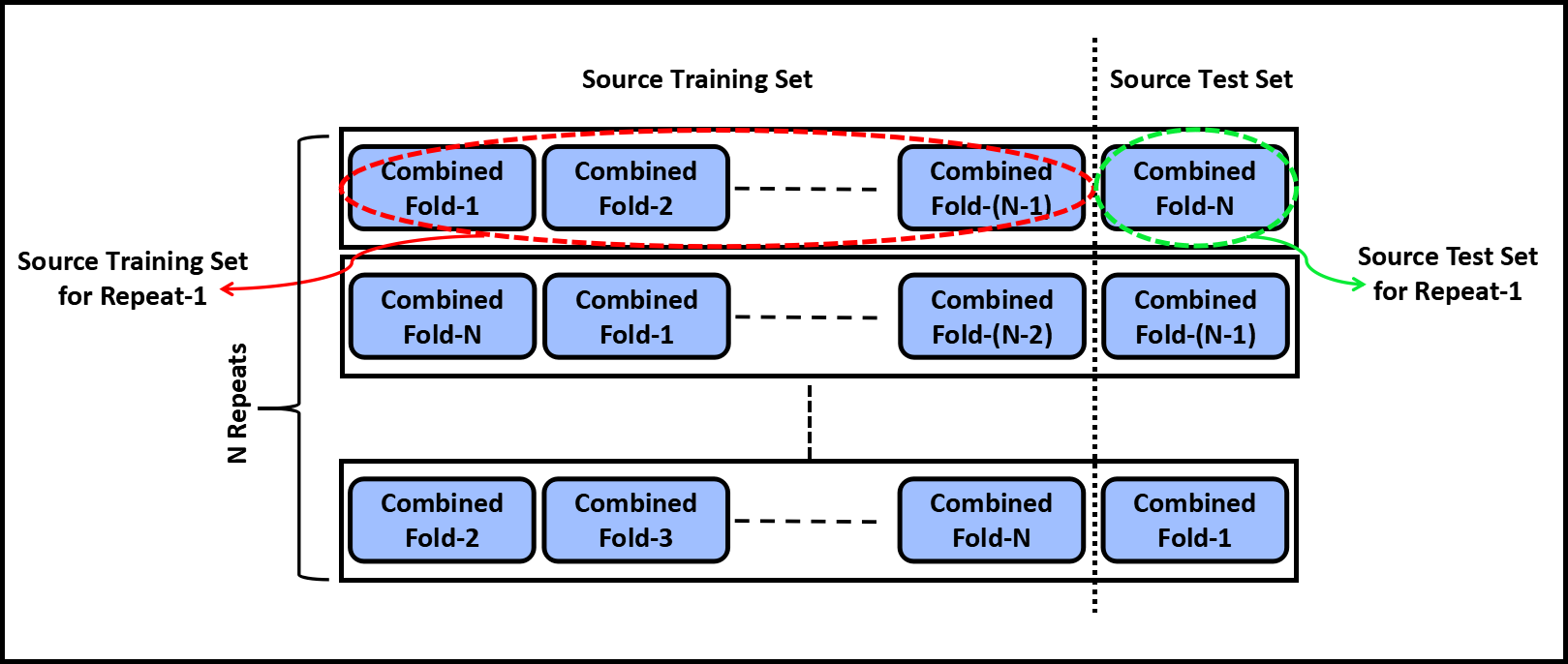}
	\caption{Set aside one combined fold to create scenes for testing and use the remaining combined folds to create scenes for training. Circular shift the folds to create source training/test set pairs for each \textit{Repeat} (total of 5 repeats).}
	\label{fig:sceneCreation2}
\end{figure*}

For each \textit{Repeat} in our experiment, we create 100 scenes for both training and testing. A 100-scene set is composed of scenes with varying number of objects (2-6 object scenes) such that it contains 20 scenes for each.

\subsection{Segmentation Solution}

In our experiments, we employ Sezgin's sketch segmentation method \cite{ref25}, which relies on dynamic programming. In order to segment a given scene, we first convert it to a collection of primitives using Sezgin's fragmentation algorithm \cite{ref26}.

Using the list of primitives, we construct a graph $G(V,E)$ in which vertices $V$ correspond to the primitives, indexed by the order in which they were drawn. The weight $w(i,j)$ associated with an edge from $v_i$ to $v_j$ in $G$ corresponds to the probability that the set of primitives between $i$ and $j$ constitutes a valid and fully-drawn symbol. The optimal segmentation $S(i,j)$ is computed through dynamic programming:

$S(i,j)=\max\begin{cases}
	$w(i,j)$\\
	{\smash{\max_{i\leq k < j}}(S(i,k).S(k,j))}
\end{cases}$\\

Sezgin computes weights $w(i,j)$ by combining an HMM and a one-class SVM such that HMM predicts class probability of the fragment groups, while one-class SVM determines whether the fragment group constitutes a valid object or not. Here, we employ a multi-class SVM to compute the weights.

\subsection{Trials}

A trial refers to the end-to-end process of active learning on a training \& test scene set pair. Throughout each trial, we measure the segmentation and recognition accuracy of the model. We initialize each trial by training the classifier (SVM in our case) with the seed set. Then, the process continues by extending the training set with the labels of the objects contained in scenes or scene segments selected by the current active learning strategy. Then, the classifier is retrained with the extended data. Pickling a scene, labeling data and retraining the classifier is called a round. Active learning continues for several rounds until a target accuracy is reached or until the user is satisfied.

We repeat our experiment 5 times with data prepared as described in Section 3.1. Hence, each \textit{Repeat} of the experiment corresponds to a trial.

Throughout our experiments, we follow the AL guidelines suggested by Yan\i k et al. We employ a multi-class SVM with an RBF kernel along with IDM features as in \cite{ref13}. We also initialize our classifier with 4 samples from each class. At each round of AL, we extend the training set by adding 3 scenes (12 objects on average) for scene-wise selection (and also for random scene selection) and 12 objects for segment-wise selection.

\section{Analysis}

To evaluate the performance of active scene learning methods, we conducted statistical analysis as suggested in \cite{ref13}. We utilize the deficiency measure \cite{ref13,ref27} to assess the relative performance of the ASL methods. The deficiency of method A with respect to B, deficiency(A,B), is a standard measure of the relative performance of the algorithms throughout the active learning process (see Figure 4). Notice that in our experiments, model accuracy is not the classification accuracy of isolated objects. Instead, model accuracy corresponds to segmentation and recognition accuracy defined as the ratio of the number of correctly segmented objects with correct labels over the total number of objects over all test scenes.

\begin{figure}[!htbp]
	\centering
	\includegraphics[height=0.25\textwidth, width=0.35\textwidth]{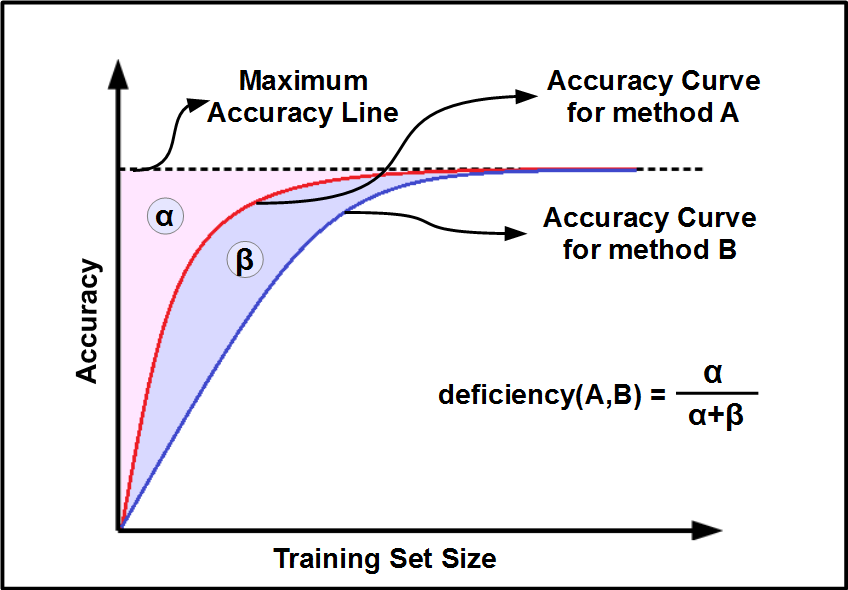}
	\caption{The deficiency is defined as the ratio of the area between the accuracy curve of method A and the maximum accuracy line; and the area between the accuracy curve of method B and the maximum accuracy line \cite{ref13}.}
	\label{fig:deficiency}
\end{figure}

In our experiments, the maximum accuracy line represents the model accuracy when it is trained with all the objects coming from the training scenes. An accuracy curve represents the sequence of accuracies (over the test set) achieved in each round of the active learning process after the model is trained with the available labeled data. Let $D=deficiency(A,B)$ be the deficiency of algorithm A computed with respect to algorithm B (see Figure 4). $D=1$ implies that the methods have a similar performance. Values less than one imply that method A is superior, while values greater than one imply that method B has superior performance.

In order to assess the statistical significance of the differences in the deficiencies of different active learning setups, we conducted multiway ANOVA tests. Throughout our analysis, we performed Mauchy's sphericity test to check whether the variances of the differences between all possible group pairs subject to ANOVA are equal. In cases where sphericity is violated, the degrees of freedom have been corrected by the Greenhouse-Geisser correction. We also performed Levene's test to check the homogeneity of variances between groups and used transformed values where appropriate. Bonferroni corrected paired t-tests were performed as Post-Hoc tests, in order to explore the mean differences across the levels of the concerned factors.

We conducted 2-way Mixed ANOVA with between group variable \textit{dataset} (with levels COAD and NicIcon) and within group variable \textit{active scene learning strategy} (with levels SwS, GM, ArM, SP, MoS). The deficiency value was taken as the dependent variable, which was computed for each active learner with respect to the random learner (values less than 1 indicate the active learning outperforming the random baseline).

\section{Results}

In this section, we present the results of our experiments. In particular, we compare performance of active scene learning methods against random selection and among each other. Moreover, we quantify the actual gains in annotation effort obtained by active scene learning methods.

\subsection{Effect of ASL strategy and dataset}

We present the results of 2-way mixed ANOVA analysis on our experiments in Table 1.

We observe a statistically significant effect of the choice of active scene learning strategy on the performance of active learning. This indicates that we should carefully select the active learning method to obtain a desired performance. We present a detailed discussion on the choice of ASL methods in the following subsections.

Another crucial observation is that dataset factor does not have a statistically significant effect on active learning performance. This is a positive result suggesting that the performance of ASL methods is consistent across these two domains. This is a good piece of news for practitioners of sketch recognition in other domains.

\begin{table}[!ht]
  \caption{F-scores and p-values for the factors of 2-way Mixed ANOVA analysis. The choice of active scene learning method has a significant effect on the performance. The choice of dataset does not have a significant effect on the performance. It indicates that performances of ASL methods are robust across the datasets.}
  \centering
    \begin{tabular}{|c|c|c|}
    \hline
    Factor & F-Score & Sig. \\
    \hline
    ASL Method & F(2.136,17.085)=10.614 & p=0.001 \\
    \hline
    Dataset & F(1,8)=0.619 & p=0.454 \\
    \hline
    \end{tabular}%
  \end{table}
\subsection{ASL vs Random Selection}

For each active scene learning method, we compute deficiency values against random selection. In Table 2, we present the estimated marginal means and (95\%) confidence intervals. To have a (confidence) upper bound less than a deficiency value of 1 indicates that active learning method confidently outperforms random selection.
\begin{table}[!ht]
  \caption{Estimated marginal means for active scene learning methods. SwS, ArM, SP and MoS methods confidently outperform random selection strategy.}
  \centering
   \resizebox{0.48\textwidth}{0.08\textwidth}
{
    \begin{tabular}{|c|c|c|c|c|}
		\hline
    \multicolumn{1}{|c|}{\multirow{2}[4]{*}{ASL Method}} & \multicolumn{1}{c|}{\multirow{2}[4]{*}{Mean}} & \multicolumn{1}{c|}{\multirow{2}[4]{*}{Std. Error}} & \multicolumn{2}{c|}{95\% Confidence Interval} \\
		\cline{4-5}
    \multicolumn{1}{|c|}{} & \multicolumn{1}{c|}{} & \multicolumn{1}{c|}{} & \multicolumn{1}{c|}{Lower Bound} & \multicolumn{1}{c|}{Upper Bound} \\
    \hline    \multicolumn{1}{|c|}{SwS} & .747  & .033  & .670  & .824 \\
    \hline
    \multicolumn{1}{|c|}{GM} & .894  & .051  & .777  & 1.011 \\
    \hline
    \multicolumn{1}{|c|}{ArM} & .833  & .049  & .720  & .946 \\
    \hline
    \multicolumn{1}{|c|}{SP} & .846  & .029  & .779  & .913 \\
    \hline
    \multicolumn{1}{|c|}{MoS} & .896  & .032  & .822  & .971 \\
    \hline
\end{tabular}%
}
  \end{table}
All active scene learning methods, except GM, outperform random selection. Another observation is that segment-wise selection has the smallest mean and upper bound values among the other. This indicates that segment-wise selection outperforms random selection much more confidently than the other active scene learning methods.

\subsection{Performance examination among ASL strategies}

Table 1 shows statistical significance in the choice of active scene learning strategy. Now, we further investigate how the performance of ASL methods contributes to this. In particular, we compare pairs of active scene learning methods to reveal their relative effectiveness based on how they fare against random selection.

In Table 3, we present the results of Bonferroni corrected paired t-tests as Post-Hoc tests. Negative mean difference and negative (confidence) upper bound indicates that there is a significant difference in performances of the two methods and the considered method (column I) confidently outperforms the other (column J).

\begin{table*}[t]
  \centering
  \caption{Bonferroni corrected paired t-test results for active scene learning methods. Having a mean difference smaller than zero (with Sig.$<$0.05) indicates that a method performs significantly better (confidently has a smaller deficiency value) than the reference method. Segment-wise selection outperforms GM, SP and MoS methods. It also tends to perform better than ArM method, but there is no statistical significance is observed in the difference between their performances.}
\newcolumntype{y}{>{\columncolor{yellow}}c}
  \resizebox{0.90\textwidth}{0.20\textwidth}
{
    \begin{tabular}{|c|c|c|c|c|c|c|c|}
    \hline
    \multicolumn{1}{|c|}{\multirow{2}[3]{*}{(I) ASL Method}} & \multicolumn{1}{c|}{\multirow{2}[3]{*}{(J) ASL Method}} & \multicolumn{1}{c|}{\multirow{2}[3]{*}{Mean Difference (I-J)}} & \multicolumn{1}{c|}{\multirow{2}[3]{*}{Std. Error}} & \multicolumn{1}{c|}{\multirow{2}[3]{*}{Sig.}} & \multicolumn{2}{c|}{95\% Confidence Interval for Difference} \\
    \cline{6-7}
    \multicolumn{1}{|c|}{} & \multicolumn{1}{c|}{} & \multicolumn{1}{c|}{} & \multicolumn{1}{c|}{} & \multicolumn{1}{c|}{} & \multicolumn{1}{c|}{Lower Bound} & \multicolumn{1}{c|}{Upper Bound} \\
    \hline
    \multicolumn{1}{|c|}{
    \multirow{4}[0]{*}{SwS}} & 
    \multicolumn{1}{c|}{GM} & -.147* & .035  & \multicolumn{1}{|y|}{.032}  & \multicolumn{1}{|y|}{-.283} & \multicolumn{1}{|y|}{-.011} \\
    \multicolumn{1}{|c|}{} & \multicolumn{1}{c|}{ArM} & -.086 & .034  & .356  & -.218 & .045 \\
    \multicolumn{1}{|c|}{} & \multicolumn{1}{c|}{SP} & -.099* & .021  & \multicolumn{1}{|y|}{.014}  & \multicolumn{1}{|y|}{-.179} & \multicolumn{1}{|y|}{-.020} \\
    \multicolumn{1}{|c|}{} & \multicolumn{1}{c|}{MoS} & -.150* & .023  & \multicolumn{1}{|y|}{.002}  & \multicolumn{1}{|y|}{-.237} & \multicolumn{1}{|y|}{-.062} \\
    \hline
    \multicolumn{1}{|c|}{\multirow{4}[0]{*}{GM}} & \multicolumn{1}{c|}{SwS} & .147* & .035  & .032  & .011  & .283 \\
    \multicolumn{1}{|c|}{} & \multicolumn{1}{c|}{ArM} & .061* & .015  & .042  & .002  & .120 \\
    \multicolumn{1}{|c|}{} & \multicolumn{1}{c|}{SP} & .048  & .026  & 1.000 & -.053 & .148 \\
    \multicolumn{1}{|c|}{} & \multicolumn{1}{c|}{MoS} & -.003 & .027  & 1.000 & -.106 & .101 \\
    \hline
    \multicolumn{1}{|c|}{\multirow{4}[0]{*}{ArM}} & \multicolumn{1}{c|}{SwS} & .086  & .034  & .356  & -.045 & .218 \\
    \multicolumn{1}{|c|}{} & \multicolumn{1}{c|}{GM} & -.061* & .015  & .042  & -.120 & -.002 \\
    \multicolumn{1}{|c|}{} & \multicolumn{1}{c|}{SP} & -.013 & .024  & 1.000 & -.104 & .078 \\
    \multicolumn{1}{|c|}{} & \multicolumn{1}{c|}{MoS} & -.063 & .031  & .775  & -.184 & .057 \\
    \hline
    \multicolumn{1}{|c|}{\multirow{4}[0]{*}{SP}} & \multicolumn{1}{c|}{SwS} & .099* & .021  & .014  & .020  & .179 \\
    \multicolumn{1}{|c|}{} & \multicolumn{1}{c|}{GM} & -.048 & .026  & 1.000 & -.148 & .053 \\
    \multicolumn{1}{|c|}{} & \multicolumn{1}{c|}{ArM} & .013  & .024  & 1.000 & -.078 & .104 \\
    \multicolumn{1}{|c|}{} & \multicolumn{1}{c|}{MoS} & -.050 & .020  & .379  & -.128 & .027 \\
    \hline
    \multicolumn{1}{|c|}{\multirow{4}[1]{*}{MoS}} & \multicolumn{1}{c|}{SwS} & .150* & .023  & .002  & .062  & .237 \\
    \multicolumn{1}{|c|}{} & \multicolumn{1}{c|}{GM} & .003  & .027  & 1.000 & -.101 & .106 \\
    \multicolumn{1}{|c|}{} & \multicolumn{1}{c|}{ArM} & .063  & .031  & .775  & -.057 & .184 \\
    \multicolumn{1}{|c|}{} & \multicolumn{1}{c|}{SP} & .050  & .020  & .379  & -.027 & .128 \\
    \hline
    \end{tabular}%
    }
  \label{tab:addlabel}%
\end{table*}%

We observe that segment-wise selection confidently outperforms all other ASL methods, except ArM. This observation is aligned with our expectation since segment-wise selection targets the most informative parts (objects) of the scenes while scene-wise selection methods select all objects in a scene.

Another observation is that ArM confidently outperforms GM. In addition, no significant difference in performance is observed between ArM and other methods (SP and MoS). Moreover, there is no significant difference among GM, MoS and SP.

Overall, segment-wise selection and ArM outperform the others, while segment-wise selection tends to perform better than ArM.

\subsection{Actual savings in the annotation effort}

So far, we shared the results of our statistical analysis. To complete the picture, we now present the actual savings achieved by ASL methods. Figure 5 illustrates the interpolated mean accuracies (over repeats) obtained with respect to the annotation effort (i.e. the number of individual objects labeled on scenes and added to the training set). In addition, the dotted horizontal line depicts the mean accuracy (over repeats) when all data is labeled.

\begin{figure*}
        \centering
        \begin{subfigure}[b]{0.48\textwidth}
                \includegraphics[width=\textwidth]{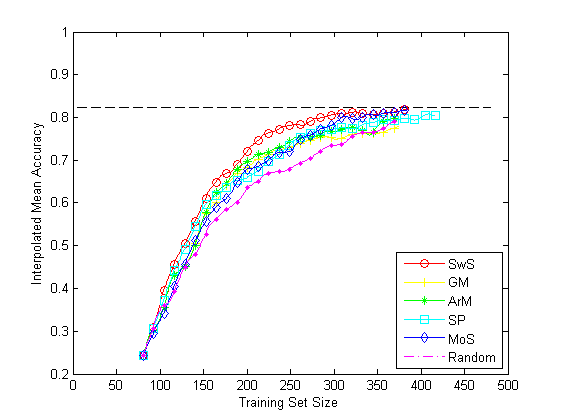}
                \caption{COAD scenes}
                \label{fig:Figure5a}
        \end{subfigure}
        \begin{subfigure}[b]{0.48\textwidth}
                \includegraphics[width=\textwidth]{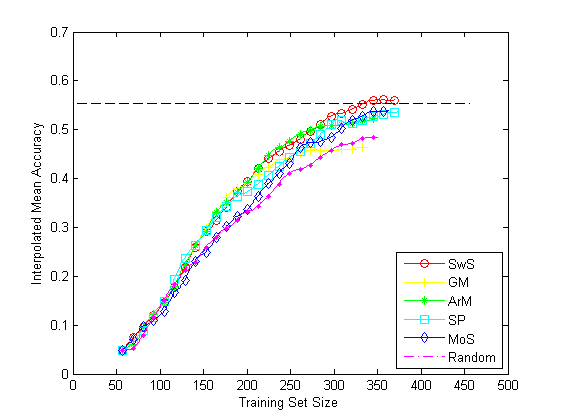}
                \caption{NicIcon scenes}
                \label{fig:Figure5b}
        \end{subfigure}
        \caption{Interpolated mean (segmentation and recognition) accuracy plots for COAD and NicIcon scenes. Active scene learning methods outperform random selection strategy. Moreover, segment-wise selection performs superior than scene-wise selection methods.}\label{fig:meanPlots}
\end{figure*}

For the COAD dataset, there are 80 seeds (4 samples from each class) and 400 unlabeled objects over 100 scenes. After initializing the model with 80 seeds, segment-wise selection achieves the top accuracy (defined as the accuracy obtained by training with all the data) by annotating 304 objects from training scenes. This is a 24\% gain in annotation effort with respect to labeling all data (400 objects). Also note that segment-wise selection gets within 1\% of the top accuracy by only labeling 40\% of the data (see Figure 5a).

For the NicIcon dataset, there are 56 seeds (4 samples from each class) and 400 unlabeled objects over 100 scenes. After initializing the model with 56 seeds, segment-wise selection obtains the top accuracy (achieved by training with all data) by annotating 277 objects from training scenes. This is a 30\% gain in annotation effort with respect to labeling all data (400 objects).

Active scene learning, specifically segment-wise selection, saves up to 30\% in annotation effort in our experiments.

\section{Discussion}

In this section, we provide a further discussion of the experiment results. This discussion will contribute to a sound understanding of the presented  active scene learning methods through a deeper analysis.

\subsection{Effect of targeted selection}

The strategy of segment-wise selection is based on targeting the most informative parts of scenes. We presented the effective performance of segment-wise selection in Section 5. Now, we extend our discussion to describe why targeting selection strategies is a desirable approach.

As segmentation model gets stronger, informativeness carried by the scenes drops since the model gets more confident in the segmented candidate objects. Confidently segmented objects contribute little if any to the performance of the model. Especially in the later rounds, finding an informative data instance turns into a hunt over all the scenes. Therefore, utilizing segment-wise selection in later rounds of AL or when the model confidence is high is definitely recommended.

In Figure 5, we observe that MoS method obtains the highest possible accuracies sooner than the other scene-wise selection methods although it falls behind them till the later rounds. This observation also supports the claim above. MoS searches for scenes which contains the most informative candidate object. In contrast, other scene-wise methods estimate scene score by considering all the candidate objects or the overall interpretation probability. As the number of informative samples decreases in the later rounds, focusing on the individual scores of the candidate objects, rather than the scene scores, contributes much more to the model's accuracy. In this sense, MoS is in the spirit of segment-wise selection due to its ability to target particularly informative object instances; but requiring additional, possibly useless, annotation effort for labeling uninformative objects that accompany the informative one in the scene.

\subsection{Effect of dataset difficulty}

Another interesting observation for segment-wise selection is that its superiority over scene-wise selection methods may be more pronounced, when the dataset is easier to learn. \cite{ref13} suggests that the NicIcon dataset is more challenging than the COAD dataset. Hence, scene sets created over these datasets allow us to compare and contrast the performance of ASL on a challenging dataset and a relatively easier one. We observe that segment-wise selection competes with scene-wise selection methods and surpasses them in the later rounds for the NicIcon data. For the COAD data, it always keeps performing above scene-wise selection methods until convergence.

It is reasonable to assume that a sample will be more representative of the others if the style variation in the dataset is smaller. Hence, such a set can be said to contain many redundant samples. Then, a targeting-based strategy such as segment-wise selection will avoid such redundant samples throughout the AL process and will achieve the top accuracies sooner. Therefore, having prior knowledge about the representativeness of the samples in the dataset might allow AL practitioners to apply active scene learning more effectively.

Another observation is that the number of classes in a dataset also affects the required amount of annotated data. When top accuracies are obtained, the average number of annotated objects (excluding seeds) per class are 15 for COAD scenes and 20 for NicIcon scenes. In other words, we obtained top accuracies by labeling 25\% fewer examples for the COAD scenes. This is in accordance with the claim \cite{ref13} that NicIcon dataset is more challenging than COAD dataset.

We also observe that segment-wise selection slightly steps above the top accuracy that can be obtained when all the data is labeled, for NicIcon scenes. We suspect that the model learned a slightly more general classifier by focusing on few but carefully selected samples.

\subsection{Effect of hard penalties}

We observe, from Table 1 and Table 2, that GM performs poorly against random selection and other ASL methods. When we investigate Figure 5, we see that it competes well against the other ASL methods and outperforms random selection till the later rounds of AL. However, in the later rounds it flattens out and stays behind even random selection. We suspect that it is due to the hard punishment of GM. Unlike ArM, it penalizes scenes drastically especially when there is at least one candidate object with very low (close to zero) informativeness. We can expect that the model is very poor in the earlier rounds and it has low confidence in the segmentation of the scenes, yielding informative candidate objects. However, as the model gets stronger through AL rounds, it will segment the scene more confidently. Even a single confidently segmented candidate object in a scene reduces the informativeness of the scene to a value close to zero. Hence, it is no surprise that GM flattens out in later rounds. This observation suggests that either we should avoid hard penalties for the scenes or we should carefully apply hard penalties and consider switching to another ASL strategy when appropriate.

\subsection{A note on tendencies for scene-length}

We observe from Figure 5 that scene-wise selection methods might have varying tendencies due to scene-length. As illustrated in Figure 5, GM tends complete the AL process (25-rounds) with fewer labeled data while SP tends to complete the process (25-rounds) with more labeled data than the other scene-wise selection methods.

For SP, as the number of (candidate) objects in the scene increases, the interpretation probability will decrease, especially with a weak model (See Section 3.2). Therefore, scenes with more objects may have a higher informativeness score (See Section 2.2.4). Then, it is expected for SP to tend to select scenes with more objects. For GM, computing the geometric mean for scenes with more objects may tend to yield smaller informativeness values, especially when the scenes contain many uninformative candidate objects. These observations indicate that adjusting informativeness scores based on the number of objects detected in a scene might be helpful especially when the scene-lengths varies too much in a dataset.

\section{Related Work}

In this paper, we presented a new active learning framework which reduces the annotation effort required to generate an accurate system for scene segmentation. We proposed two selection schemes under the ASL framework and investigated their variants. Then, we demonstrated their effective performance by analyzing how they fare against random selection and among each other.

There are several custom interfaces \cite{ref8,ref9,ref10,ref11} for labeling sketch data in the literature. Although the user friendly interfaces ease the annotation process, they do not consider reducing the number of samples to be labeled. 

Work by Plimmer et al. \cite{ref12} proposes to use auto-labeling to automatically annotate unlabeled samples. They use a classifier trained with partially labeled dataset to automatically annotate unlabeled samples.  Unfortunately, it is known that not all unlabeled examples are equally useful (some carry redundant information, whereas others are more informative). This method lacks a mechanism for deciding which samples should be labeled first to gain maximum benefit. This leads to suboptimal use of the valuable annotation resources. Furthermore, without a sufficiently accurate classifier, auto-annotation is prone to yield incorrect labels, hence a manual label validation and correction effort is required.

Yan\i k and Sezgin \cite{ref13} demonstrate effective use of active learning for sketch recognition. Although they conduct a rigorous statistical analysis to investigate factors affecting AL performance, and provide the community valuable insight on the use of AL, their work focuses on applying AL on isolated sketch recognition rather than scenes. Hence, our work fills this niche and enables practitioners to use AL on scene data.

There are lines of work utilizing active learning for image segmentation including domains such as CT-Scans \cite{ref14,ref15,ref16}, 3D images \cite{ref15,ref16} and hyper/multi-spectral images \cite{ref17}. While some of these \cite{ref14,ref15,ref16,ref17,ref18} apply AL on semantic segmentation, others \cite{ref19,ref20,ref21,ref22} focus on foreground extraction (object segmentation). However, these approaches compute informativeness at fine levels of granularity such as pixels, super-pixels, regions, voxels \cite{ref15}, (3D) slices \cite{ref16}, and tokens \cite{ref23,ref24}. These approaches advocate a primitive-based labeling strategy (as opposed the holistic approach we take). Rather than identifying whole objects, and assigning labels to all primitives of an object at once, labeling is performed over individual super-pixels, tokens or primitives that constitute the scene. 

The superpixel level approaches have two serious limitations. First, they assume that individual super-pixels carry substantial information to allow reasonable superpixel level classification. This assumption may hold in limited scenarios such as background segmentation where the foreground and background super-pixels have distinct intensity, color and texture statistics to allow training superpixel level classifiers. For sparse inputs, as in the case of sketches, primitives (such as stroke fragments or geometric primitives) are highly generic, and primitive-level prediction of object classes has proved impractical. Hence, we advocate training object-level classifiers, and perform active scene segmentation at the object granularity.

Second, from a user interaction point of view, annotation at the granularity of super-pixels is highly impractical. A single object may consist of hundreds of super-pixels. Active learning at this granularity is bound to require far more annotator interventions compared to the holistic case where actions are taken at the object level. For example, [18] shows that reaching 97\% of the performance of the fully supervised accuracy requires 17\% of the super-pixels to be annotated. At the 17\% ratio, one might as well label the entire dataset without any active learning, since it would actually require fewer annotation actions, unless one is operating in a domain where objects have fewer than 6 super-pixels on average. For computer vision problems, oversegmentation without leakage (undersegmentation error \cite{ref28}) is bound to require orders of magnitude more super-pixels, which makes pixel-based active learning more costly than plain annotation in practice.  

There is also some work applying AL on sequence labeling \cite{ref23} and structured output spaces \cite{ref24}. Again, these systems operate at the granularity of primitives (i.e. tokens), which are readily available in the form of words.

In other lines of work, Vijayanarasimhan et al. \cite{ref29} and Settles et al. \cite{ref30,ref31} apply AL on multi-instance learning (MIL) task for image and text categorization. Although the bags in MIL contain more than a single object (e.g. search engine results) or an object with smaller parts (e.g. image segments and text passages), there is no notion of building a model targeting segmentation. Their aim is to figure out how much a sample contributes to the label (decision) of the bag. Unlike applying AL on MIL, we focus on building an accurate segmentation and recognition model for scenes containing arbitrary number of objects in arbitrary configurations. Moreover, we show that our framework can reduce the amount of data to be labeled drastically and can still reach top accuracy figures.

\section{Future Work}

In this paper, we presented an active scene learning framework to enable AL on scene data. We propose segment-wise selection and scene-wise selection. Future work may contribute new selection schemes and informativeness measures under Active Scene Learning framework in several ways. Introducing batch selection strategies may improve both segment-wise and scene-wise selection schemes. For scene-wise selection, weighting several informativeness measures might overcome drawbacks of individually computing these informativeness measures, especially with a dynamic weight update strategy. Furthermore, a mechanism for dynamically updating weights or switching among informativeness measures require a comprehensive investigation of a variety of factors.

We observed that ASL methods may have tendencies to favor scenes with more/fewer objects. Investigating extreme cases when too many or too few objects exist in all scenes might provide a better understanding of whether such tendency has a strong effect on performances of the ASL methods. In addition, proposing strategies for adjusting informativeness scores based on the number of (candidate) objects in the scene may make ASL methods more robust to variations in the scene length.

We utilized a dynamic programming based segmentation solution in our experiments. It would be interesting and complementary to apply ASL methods along with various other segmentation methods as well.

\section{Summary}

We presented the Active Scene Learning framework to enable use of active learning on unsegmented data. In particular, we investigated whether candidate objects produced by segmentation carry sufficient information to benefit from applying active learning. In the course of doing so, we proposed 2 selection schemes (namely segment-wise selection and scene-wise selection), then we demonstrated the usability and the effectiveness of the ASL framework through an empirical analysis as suggested in the literature \cite{ref13}. We observed that ASL methods confidently outperform random selection, and also obtain up to 30\% gain in annotation effort to achieve top accuracy figures. By filling the niche in the active learning literature, ASL framework allows application of AL on a broader range of applications.

We suggested several basic informativeness measures under scene-wise selection scheme. Then, we examined performances of these scene-wise selection methods along with performances of segment-wise selection and random selection strategies via a statistical analysis followed by a detailed discussion. We demonstrated that targeting specific parts of scenes during the active learning process yields superior performance. Moreover, we showed that segment-wise selection is a preferable selection strategy in comparison to scene-wise selection methods and random selection due to its ability to target a specific part of a scene.

We provided a detailed discussion of our experimental results, which we hope, will serve as a valuable guide for active learning practitioners. Our work draws attention to several important factors which may have effect on ASL methods; such as (style) variation in data, hard penalties for computing scene informativeness and tendencies towards scenes with more/fewer objects. We foresee that such a detailed discussion along with the results of our statistical analysis will serve as guidelines for the community members as they design, apply and evaluate new ASL methods.

\section{Conflict of Interest} None to declare.

\bibliographystyle{model3-num-names}
\bibliography{references}

\begin{*}
\newline
\textbf{Erelcan Yan\i k} received his bachelor's degree from Ko\c c University in 2011. He completed his master's degree in Intelligent User Interfaces Laboratory at Ko\c c University. He is currently working in the industry as a software/machine learning engineer. His primary area of work is to design and build machine learning models/modules (especially for NLP and CV applications) that are applicable to big data problems. He also designs and develops big data architectures and pipelines for managing knowledge graphs, in order to create more value out of the information extracted from integrated (ML) modules.
\end{*}

\begin{*}
\newline
\textbf{T. Metin Sezgin} graduated summa–cum laude with honor's from Syracuse University in 1999. He received the MS and PhD degrees from the Massachusetts Institute of Technology in 2001 and 2006. He subsequently joined the University of Cambridge as a postdoctoral research associate, and held a visiting researcher position at Harvard University in 2010. He is currently an associate professor at Koc University, Istanbul and directs the Intelligent User Interfaces (IUI) Laboratory. His research interests include intelligent human computer interfaces and HCI applications of machine learning. He is particularly interested in applications of these technologies in building intelligent pen–based interfaces.
\end{*}

\end{document}